\documentclass[letterpaper, 10 pt, conference]{ieeeconf}  
\IEEEoverridecommandlockouts

\usepackage{cite}
\usepackage{amsmath,amssymb,amsfonts}
\usepackage{algorithmic}

\usepackage{xcolor}
\usepackage{diagbox}
\usepackage{caption,subcaption}
\usepackage{pifont} % \ding{55} script X:  http://ctan.org/pkg/pifont

\usepackage{graphicx}
\usepackage{times} 
\usepackage[font=small,skip=3pt,tableposition=top]{caption}
\usepackage[bookmarks=true]{hyperref}

\definecolor{darkgreen}{RGB}{0,127,0}
\definecolor{darkred}{RGB}{200,0,0}

\def\greencheckmark{\textcolor{darkgreen}{\checkmark}}
\def\redxmark{\textcolor{darkred}{\ding{55}}}  % pifont

\newcommand{\norm}[1]{\left\lVert#1\right\rVert}
\newcommand{\R}{\mathbb{R}}  % real numbers
\begin{document}

%\title{PredictionNet}
\title{\LARGE \bf PredictionNet: Real-Time Joint Probabilistic Traffic Prediction \\ for Planning, Control, and Simulation}
\author{Alexey Kamenev, Lirui Wang, Ollin Boer Bohan, Ishwar Kulkarni, Bilal Kartal \\ Artem Molchanov, Stan Birchfield, David Nist{\'e}r, Nikolai Smolyanskiy  \\
NVIDIA % <-this % stops a space
% \thanks{All authors are affiliated with NVIDIA Corporation.}%
}

% \author{\IEEEauthorblockN{Authors}
% \IEEEauthorblockA{\textit{NVIDIA} \\
% email@nvidia.com}
% }

\maketitle

%Predicting the future motion of traffic agents is crucial for safe and efficient autonomous driving. To this end, we present PredictionNet, a deep neural network (DNN) that predicts the motion of all surrounding traffic agents together with the ego-vehicle's motion. All predictions are probabilistic and are represented in a simple top-down rasterization that allows an arbitrary number of agents. Conditioned on a multi-layer map with lane information, the network outputs future positions, velocities, and backtrace vectors jointly for all agents including the ego-vehicle in a single pass. Trajectories are then extracted from the output.  The network can be used to simulate realistic traffic, and it produces competitive results on popular benchmarks.  More importantly, it has been used to successfully control a real-world vehicle for hundreds of kilometers, by combining it with a motion planning/control subsystem. The network runs faster than real-time on an embedded GPU, and the system shows good generalization (across sensory modalities and locations) due to the choice of input representation. Furthermore, we demonstrate that by extending the DNN with reinforcement learning (RL), it can better handle rare or unsafe events like aggressive maneuvers and crashes.

\begin{abstract}
Predicting the future motion of traffic agents is crucial for safe and efficient autonomous driving. To this end, we present PredictionNet, a deep neural network (DNN) that predicts the motion of all surrounding traffic agents together with the ego-vehicle's motion. All predictions are probabilistic and are represented in a simple top-down rasterization that allows an arbitrary number of agents. Conditioned on a multi-layer map with lane information, the network outputs future positions, velocities, and backtrace vectors jointly for all agents including the ego-vehicle in a single pass. Trajectories are then extracted from the output.  The network can be used to simulate realistic traffic, and it produces competitive results on popular benchmarks.  More importantly, it has been used to successfully control a real-world vehicle for hundreds of kilometers, by combining it with a motion planning/control subsystem. The network runs faster than real-time on an embedded GPU, and the system shows good generalization (across sensory modalities and locations) due to the choice of input representation. Furthermore, we demonstrate that by extending the DNN with reinforcement learning (RL), it can better handle rare or unsafe events like aggressive maneuvers and crashes.\footnote{Video at \url{https://youtu.be/C7Nb3DRjFP0} .}
\end{abstract}

%\begin{IEEEkeywords}
%DNN, Autonomous Driving, prediction
%\end{IEEEkeywords}

\section{INTRODUCTION}

Safe and effective autonomous driving requires accurately predicting the motion of nearby actors, such as vehicles, pedestrians, and cyclists. An autonomous vehicle needs to anticipate whether a nearby car will keep moving forward at constant speed, brake suddenly, or cut into an adjacent lane. Predicting other agents' behaviors has been shown beneficial both in adversarial and cooperative domains~\cite{yoshida2008game,hernandez2019agent}.

Brute-force enumeration of all possible actions leads to combinatorial explosion, creating a computational bottleneck.  Similarly, most recent approaches  \cite{lee2017desire}, \cite{rhinehart2019precog}, \cite{chai2019multipath}, \cite{djuric2020multixnet}, \cite{casas2020implicit}, \cite{salzmann2020trajectron++} use very deep networks (e.g., ResNet50), predict each agent separately or use batches of cropped feature maps, which precludes real-time operation. We seek a system that jointly predicts the motions of all actors, is not limited to the number of actors, is agnostic to input sensory modality, runs in real-time, and handles rare/unsafe events.

\begin{figure}
    \centering
\begin{subfigure}{1.0\linewidth}
 \includegraphics[width=1\linewidth]{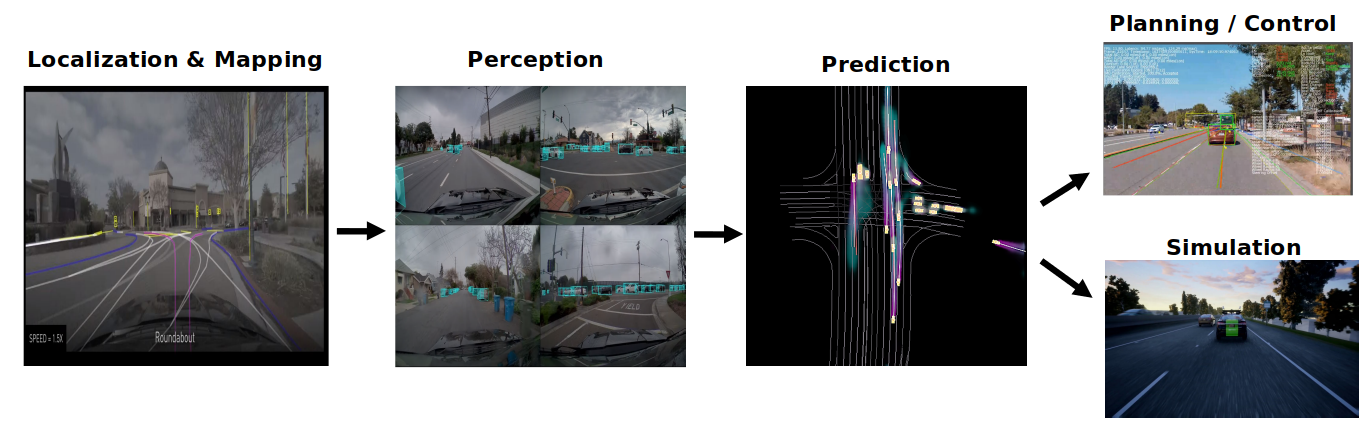}
 \label{fig:framework_diagram}
\end{subfigure}
\vspace{-2mm}
\caption{Simplified autonomous driving diagram. PredictionNet can be used for planning/control, or simulation.}
\vspace{-4mm}
\label{fig:frameworkoverview}
\end{figure}

In this paper, we propose a novel system which, to our knowledge, is the first to satisfy these requirements. Our system, called PredictionNet, relies on a simple yet flexible representation for both input and output. The current state (input) is a top-down multi-channel rasterization of the perceived vehicle boundaries and relevant scene context provided by the map's lane dividers. The method is agnostic to the perception modality as it relies on its own representation created by top-down rasterization. The predicted state (output) is a top-down representation providing the probability density function (PDF) of all traffic actors' future poses along with estimated velocities and backtrace vectors to aid trajectory extraction. The system is trained on real driving data to simultaneously predict full PDFs of the motion of all observed traffic agents and the ego-vehicle in a single pass.
The network can be used for both realistic simulation and closed-loop planning/control (Fig.~\ref{fig:frameworkoverview}).

We have incorporated PredictionNet into the closed-loop planning/control stack of a real-world autonomous driving system. The network runs \emph{faster than real-time}, providing sufficient time for the planner to react safely.
%\footnote{ Our model inference runs at \textbf{5~ms} on an embedded NVIDIA Drive AGX GPU and \textbf{4~ms} on Titan Turing GPU plus we need \textbf{3~ms} for pre-/post-processing.  By comparison, our closest competitor MultiPath~\cite{chai2019multipath} is an order of magnitude slower.}
The network effectively predicts on camera and radar data, despite training only on LiDAR data. We show improvements in real driving behavior compared to an analytical baseline, as well as competitive offline results on both the INTERACTION prediction benchmark~\cite{zhan2019interaction} and an internal dataset.

The learned network can also be used for real-time traffic simulation. As such, it avoids the common problems of simulators that rely on either hand-coded rules or replay from driving logs. The former cannot capture complex interactions, while the latter cannot be used to study rare or unsafe behaviors, because datasets collected from actual drives necessarily under-represent such events. We use this simulator, together with reinforcement learning (RL), to extend our network with an ego-policy head that encourages collision-free driving beyond the expert dataset distribution.

Our contributions are as follows:
\begin{itemize}
    \item A deep neural network (DNN) that predicts all traffic agents, including the ego-vehicle, and models their interactions in one pass; the DNN consists of a 2D convolutional neural network (CNN) encoder/decoder and two 2D recursive neural networks (RNNs);
    \item Flexible probabilistic output representation that captures an arbitrary number of traffic agents;
    \item Faster than real-time performance, and agnostic to input sensory modalities (LiDAR / radar / camera);
    \item A model-based reinforcement learning extension to handle rare or unsafe events.
\end{itemize}

\section{PREVIOUS WORK}

\textbf{Traffic Prediction.}
A number of analytical and deep learning approaches to vehicle prediction have been proposed. Analytical methods use heuristics such as lane following with constant velocity. Despite the stability of these methods, DNNs often provide better predictions across diverse traffic scenarios. Specifically, DESIRE \cite{lee2017desire} uses a conditional variational auto-encoder (CVAE \cite{kingma2013auto}) to get future prediction samples, then uses an RNN for ranking. A top-down LiDAR scan is encoded by a CNN to provide a scene context. PRECOG~\cite{rhinehart2019precog} uses similar input encodings and employs a probabilistic model of future interactions between agents, conditioned on the goal of the ego-actor.  MultiPath~\cite{chai2019multipath} uses a set of anchors for prediction. Each anchor is a mode in a trajectory distribution and together they form a Gaussian mixture model. VectorNet~\cite{gao2020vectornet} relies on vectorization of map lanes and past trajectories via a graph neural network (GNN \cite{wu2020comprehensive}) and uses attention mechanism for predicting. MultiXNet~\cite{djuric2020multixnet} provides object detection jointly with trajectory prediction from LiDAR input. The DNN uses a batch of cropped top-down feature maps encoded by the detector stage to produce multi-modal predictions. ILVM~\cite{casas2020implicit} defines the distribution over future trajectories by using implicit latent variable model. The system uses a CNN to encode a scene, crops its feature maps for each actor and then uses GNN for latent learning and prediction. Trajectron++ \cite{salzmann2020trajectron++} jointly learns to predict controls for all agents with a GNN and then integrates agent trajectories using vehicle-dynamics models. Most DNN-based approaches predict behavior for each traffic agent separately and only predict behavior for non-ego agents. Therefore, they cannot be used in ego-motion planning. Most systems are too slow for real-time applications. 

\textbf{Traffic Simulation.}
Simulation is a safe and controllable way to evaluate system performance against diverse scenarios. Rule-based simulators such as  CARLA  \cite{dosovitskiy2017carla} and SUMO \cite{lopez2018microscopic} are widely used in benchmarks and tests. However, policies learned in these simulators often transfer poorly to reality due to lack of realistic behaviors. Several recent works~\cite{suo2021trafficsim,bergamini2021simnet} have been proposed to simulate traffic in a data-driven fashion. TrafficSim \cite{suo2021trafficsim} uses a GNN and a generative model to synthesize rare scenarios. However, the model runs at $0.5$ frames per second (FPS) and therefore hard to use in real-time driving. SimNet \cite{bergamini2021simnet} trains a single-step action model for realistic simulation to evaluate motion planners.

\textbf{Planning and Control.}
There is a large body of literature on the use of traffic predictions for online planning and offline policy learning. Analytical traffic forecasting and hand-coded cost functions~\cite{buehler2009darpa,fan2018baidu} are used to satisfy constraints such as traffic rules and collision avoidance. On the other hand, learning-based planning methods \cite{zeng2019end,sadat2019jointly,henaff2019model} have shown progress on reactive behavior and generalization in complex driving scenarios. Many works \cite{pomerleau1989alvinn,ratliff2006maximum,bansal2018chauffeurnet,codevilla2018end} use imitation learning (IL) to mimic driving behaviors. The simplest example is behavior cloning (BC) that suffers from the known distribution-shift problem \cite{ross2011reduction}. Another paradigm is reinforcement learning (RL). Due to the high-dimensional inputs and sparse rewards, model-free RL \cite{wolf2017learning,kiran2021deep,shalev2016safe,lillicrap2015continuous,chen2019model,saxena2020driving} approaches are sample inefficient, rendering them impractical for real-world driving. Using model-based RL with learned traffic models allows us to train realistic policies that can learn to react to rare/unsafe events injected in a simulator.

\section{METHOD}

\subsection{PredictionNet}
\begin{figure}
    \centering
\begin{subfigure}{1\linewidth}
 \includegraphics[width=1\linewidth]{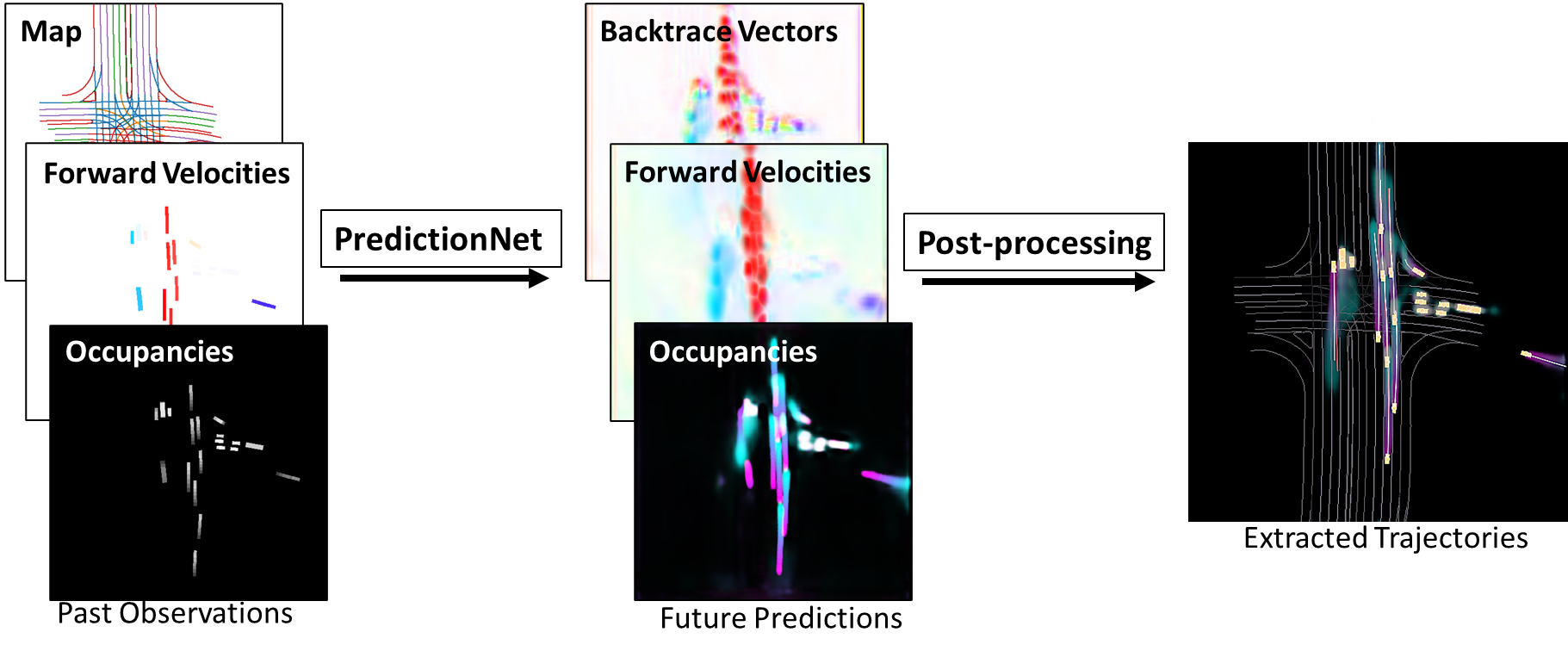}
 \label{fig:pipeline_diagram}
\end{subfigure}
\caption{Pipeline diagram. The velocities and vectors are shown using a standard optical flow color map, while the occupancy and trajectory visualizations transition from magenta to cyan in time. The yellow boxes in the far right image are vehicles.} 
% History observations (including map, occupancies, and velocities) are fed into the PredictionNet to infer the future occupancies and velocities. The raw output can be used to extract trajectories in the post-processing step.  }
\vspace{-6mm}
\label{fig:pipeline}
\end{figure}

\begin{figure*}
    \centering
\begin{subfigure}{0.8\linewidth}
 \includegraphics[width=1\linewidth]{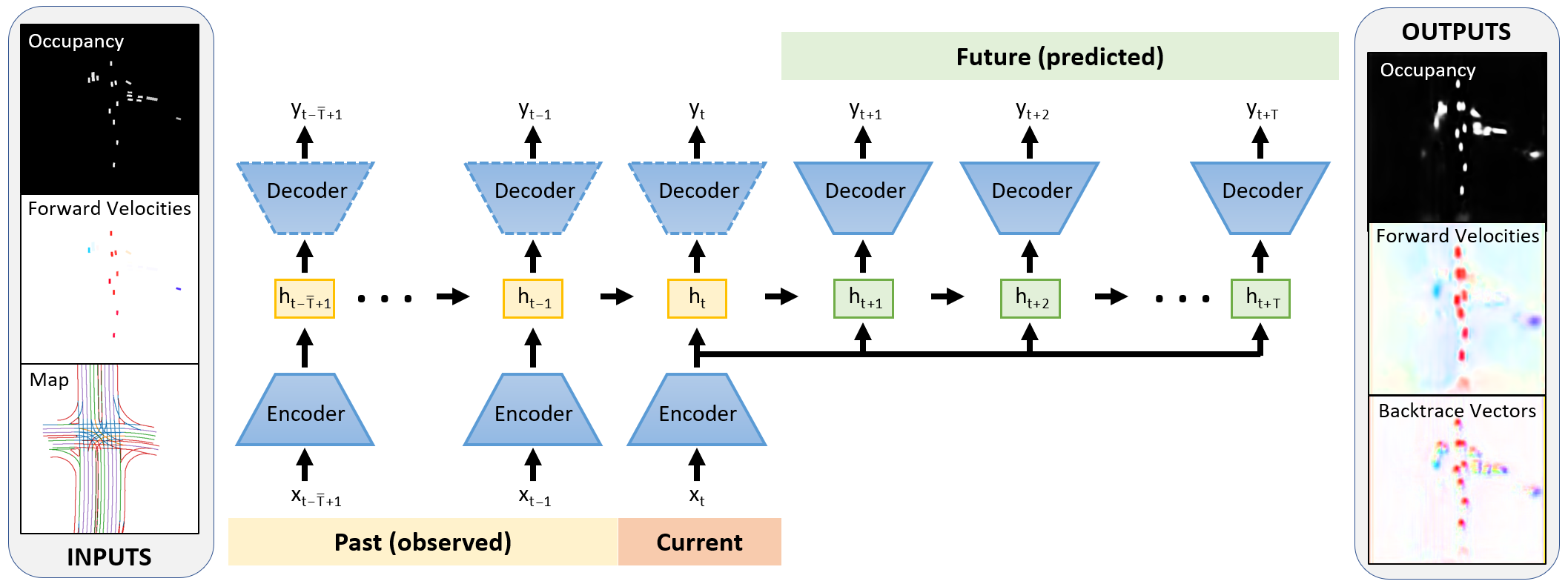}
 \label{fig:network_diagram}
\end{subfigure}
\caption{Network diagram. Our network uses CNN encoders/decoders and two recursive neural networks (RNN)---one for processing past-present observations, and another for predicting the future.  
The past-present RNN (yellow) encodes the observed velocities and occupancy along with the map. The latent state is then rolled out by the future RNN (green) to estimate future velocities and occupancy, along with backtrace vectors. The future RNN cells get encoded present input as context. The resulting sequence of future frames is used to extract the trajectories of all vehicles. The past decoders (dashed) are only used for training.}
\label{fig:PredictionNet_archxx}
\end{figure*}

\subsubsection{Data representation}

% Let $M$ be an $m \times n$ rasterized top-down map.
% Let $S_t$ is the input at time $t$.
% $S_t$ contains the map $M$, 

Our approach to traffic prediction relies upon a single deep neural network to transform the current traffic scene input into predicted future motion for all traffic agents, including the ego-vehicle; and a post-processing step to extract trajectories (Fig.~\ref{fig:pipeline}). 
The traffic scene, which is rasterized in a top-down view, is perceived by a separate system (using camera, radar, and/or LiDAR) to include data for all traffic agents surrounding the ego-vehicle along with the current map (e.g., lane dividers).
This approach allows the network to predict the motion of an arbitrary number of traffic agents without any extra compute.  

Let $M_t \in \R^{2 \times h \times w}$ be a rasterized representation of the road geometry, transformed into the ego-vehicle's frame of reference at the time $t$ of prediction.  The ego-vehicle is at the center, oriented to point upward.
Each entry in $M_t$ indicates one of a discrete number of road line types, depending upon the travel direction allowed, or zero if there is no road line at that pixel; the second channel stores the altitude of each lane divider relative to the ego-vehicle. Similarly, let $O_t \in \R^{h \times w}$ be a binary image indicating the occupancy map of vehicles, where the vehicles are rendered as oriented rectangles.  

Let $V_t \in \R^{2 \times h \times w}$ represent the rasterized 2D forward velocities of those vehicles on the ground plane.  
Note that $O_t(\cdot,\cdot)=0$ implies $V_t(\cdot,\cdot)=0$, since only the pixels corresponding to vehicles have non-zero velocities. 
The dynamic inputs $O_{t}$ and $V_{t}$ are stacked into a tensor $x^{\text{dynamic}}_t \in \R^{3 \times h \times w}$, where the 3 channels arise from stacking the single-channel occupancy with the two-channel velocity images.
The static input is simply $x^{\text{static}}_t = M_{t} \in \R^{2 \times h \times w}$, where the 2 channels represent the lane divider types and altitude.

% For notational simplicitly, let us treat $t$ as an index.
% The dynamic inputs $O_{t-{\overline T}+1\,:\,t}$ and $V_{t-{\overline T}+1\,:\,t}$ for the past-present ${\overline T}$ frames are stacked into a tensor $x^{\text{dynamic}}_t \in \R^{{\overline T} \times 3 \times h \times w}$, where the 3 channels arise from stacking the single-channel occupancy with the two-channel velocity images.
% The static input is simply $x^{\text{static}}_t = M_{t} \in \R^{2 \times h \times w}$, where the 2 channels represent the lane divider types and altitude.
% The input to the network at time $t$ is the state $x_t = (x^{\text{static}}_t, x^{\text{dynamic}}_t)$. The input tensor for each timestep is processed by an identical encoder. The static input is processed by convolutional layers and added to all processed dynamic inputs for past-present times (see Table~\ref{tab:netarch}).

The output is $y_t = ({\widehat O}_{t+1}, {\widehat V}_{t+1}, {\widehat W}_{t+1})$, where ${\widehat O}_{t+1} \in \R^{h \times w}$ is the predicted occupancy map for the next timestep, ${\widehat V}_{t+1} \in \R^{2 \times h \times w}$ contains the rasterized predicted forward velocities, and ${\widehat W}_{t+1} \in \R^{2 \times h \times w}$ contains additional backtrace vectors used to recover trajectories.
A pixel in ${\widehat V}_{t+1}$ specifies the predicted \emph{tangential} forward velocity from $t$ to $t+1$ in meters per second. A pixel in ${\widehat W}_{t+1}$ specifies the estimated vector from that pixel's location at $t+1$ to the \emph{center} of the corresponding vehicle at the previous time $t$.  (${\widehat W}$ captures the actual path, not the tangential motion.)

\subsubsection{Network Architecture and Training}

Let ${\overline T}$ be the history buffer length, and let $T$ be the maximum horizon.
To summarize the past-present ${\overline T}$ observations, and to predict $T$ timesteps into the future with a single forward pass, the network architecture is a simple 2D CNN encoder-decoder with two RNNs: one for processing past-present inputs, and another for predicting the future.
See Fig.~\ref{fig:PredictionNet_archxx}. 

For notational simplicitly, let us treat $t$ as an index.
The input to the CNN encoder at time $t$ and offset $k$ is the state $x_{t-k} = (x^{\text{static}}_t, x^{\text{dynamic}}_{t-k})$, where $k=0,\ldots,{\overline T}-1$. 
That is, the input dynamic tensor for each of ${\overline T}$ timesteps is processed by an identical encoder. Likewise, the static input for the current time $t$ is processed and added to all processed dynamic inputs for the ${\overline T}$ past-present timesteps.  See Table~\ref{tab:netarch} for details.

The encoded output is the latent state $h_{t-k}$.
For the past-present timesteps $k=0,\ldots,{\overline T}-1$, the first RNN transforms $h_{t-k}$ into $h_{t-k+1}$ using $x_{t-k}$.
But we do not have access to future observations, i.e., $x_{t+k}$, where $k>0$.
Instead, we note that $h_t$ summarizes all of the ${\overline T}$ past-present observations, and thus it can be used for future predictions. 
Thus, for each timestep $t+k$ after the current time, the encoded $x_t$ (that is, the encoded input for the present time) is provided as input context to the second RNN, which outputs $h_{t+k+1}$. 
Future prediction is unrolled in an open-loop fashion where only the RNN hidden states are passed between timesteps.

%and we use $x_t$ (encoded input for present time) as input context for RNN cells.

% The dynamic inputs $O_{t-{\bar T}+1\,:\,t}$ and $V_{t-{\bar T}+1\,:\,t}$ for the past ${\bar T}$ frames are stacked into an input $x^{\text{dynamic}}_t \in \R^{{\bar T} \times 3 \times h \times w}$, where the 3 channels arise from stacking the single-channel occupancy with the two-channel velocity images.
% The static input is simply $x^{\text{static}}_t = M_{t} \in \R^{1 \times 2 \times h \times w}$, where the 2 channels represent the type and altitude of nearby lane dividers.
% The inputs are top-down scene images and are encoded shared CNN encoders.
% % With slight abuse of notation, let us represent the combined input as $x_t = (x^{\text{static}}_t, x^{\text{dynamic}}_t)$.
% For past-present observations up to the current time $t$, the internal past RNN state $h_{t-k-1}$ is updated to $h_{t-k}$ using $x_{t-k}$, where $k=0, \ldots,  {\overline T}-1$, and ${\overline T}$ is the history buffer length.  
% Given that $h_t$ summarizes past-present observations, it can be used for future predictions. 
% Thus, for timestep $t+k$ after the current timestep, since we do not have full observations, we provide $x_t$ as input context and update $h_{t+k}$ iteratively for $k=1, \ldots,T$ up to the maximum horizon $t+T$. 
% The future prediction is unrolled in an open-loop fashion where only the RNN hidden states are passed between timesteps and we use $x_t$ (encoded input for present time) as input context for RNN cells.

The network is trained end-to-end with supervised learning. The latent state $h_{t+k}$ at each timestep is decoded into $y_{t+k}$, where $k=1-{\overline T},\ldots,T$. We use a combination of focal loss \cite{lin2017focal} for the occupancy and $L_2$ loss for the predicted backtrace vectors and velocities:
\begin{align}
    L(y_t) =  &\lambda_0 \,L_{\text{focal}} \left( O_{t}, \widehat{O}_{t} \right) +  \nonumber \\&\lambda_1\norm{V_{t}-\widehat{V}_{t}}^2_2 +\lambda_2\norm{W_{t}-\widehat{W}_{t}}^2_2, 
\end{align}
where $\lambda_0,\lambda_1,\lambda_2$ are scalar weights.

\subsubsection{Post-processing}

The DNN outputs (occupancy, velocity, backtrace vectors) provide a general, non-parametric prediction of future traffic. When discrete trajectories are required, we compute trajectories via post-processing of the raw network outputs. Trajectories begin at the initial position/velocity for each agent and are iteratively rolled out through Euler integration of the DNN's predicted velocity tensor. At each iteration, time is advanced by a fixed increment, position is advanced by a linear step using the current velocity, and a new velocity is sampled from the DNN's predicted velocity tensor at the resultant time/position. Additionally, we apply two modifications:

1. \emph{Linear fallback}: To avoid sampling invalid velocities from regions with no predicted occupancy, we reuse the previous timestep's velocity if predicted occupancy is low.

2. \emph{Drift correction}: Accumulation of small errors can cause trajectories to drift around curves, such that the backtrace vector becomes misaligned with the forward velocity vector. To avoid these errors, we apply a small correction step after each linear step, proportional to the difference between the forward velocity vector and the reversed backtrace vector.

Trajectory post-processing uses a small number of parameters set empirically on the training set after the network is trained. Post-processing takes $T$ constant timesteps to complete and runs in parallel for all agents on the GPU.

% Older description that has the recurrence, but not the explanation for why it exists:

% Trajectory extraction for all vehicles including the ego-vehicle is done via post-processing of the output tensors $\{y_t,y_{t+1},...,y_{t+T}\}$. 
% We start each trajectory at the initial position and velocity ($p_t, v_t$) given by the perception system. We roll out subsequent positions and velocities ($p_{t+k}, v_{t+k}$) by integrating over the predicted velocity tensor, using the predicted occupancy tensor as a confidence signal, and using the backtrace vectors for drift-correction. This roll-out uses the following recurrence: 
% \begin{align}
%     p_{t+1} &= p_t + v_t \Delta t\\
%     \alpha_{t+1} &= \sigma(w_{\alpha}\cdot \widehat O_{t+1}[p_{t+1}] + b_{\alpha})\\
%     p_{t+1} &= p_{t+1} + \alpha_{t+1} \cdot
%     \emph{corr}_{p}\left(\widehat V_{t+1}[p_{t+1}], \widehat W_{t+1}[p_{t+1}]\right) \\
%     v'_{t+1} &= \widehat V_{t+1}[p_{t+1}] + \alpha_{t+1} \cdot \emph{corr}_{v}\left(\widehat V_{t+1}[p_{t+1}], \widehat W_{t+1}[p_{t+1}]\right)\\
%      v_{t+1} &= \alpha_{t+1}\cdot v'_{t+1} + (1 - \alpha_{t+1}) \cdot v_t
% \end{align}
% where $\sigma$ is the logistic sigmoid function; and $w_{\alpha}, b_{\alpha} \in \mathbb R$, $\emph{corr}_{p}, \emph{corr}_{v} \in \mathbb R^{2\times4}$ are parameters set empirically (after PredictionNet training has completed) to minimize average displacement error (ADE) on the training set. 
% 

\subsection{PredictionNet for Closed-loop Traffic Simulation}

% \begin{figure}
%     \centering
% \begin{subfigure}{0.85\linewidth}
%  \includegraphics[width=1\linewidth]{fig/latent_samples.png}
% \end{subfigure}%
% \caption{Samples of latent vectors in PredictionNet. The latent space captures past-present agent interactions and map geometry.}
% \label{fig:Prednet_latent}
% %\vspace{-4mm}
% \end{figure}

\begin{figure}
    \centering
\begin{subfigure}{1.0\linewidth}
 \includegraphics[width=1\linewidth]{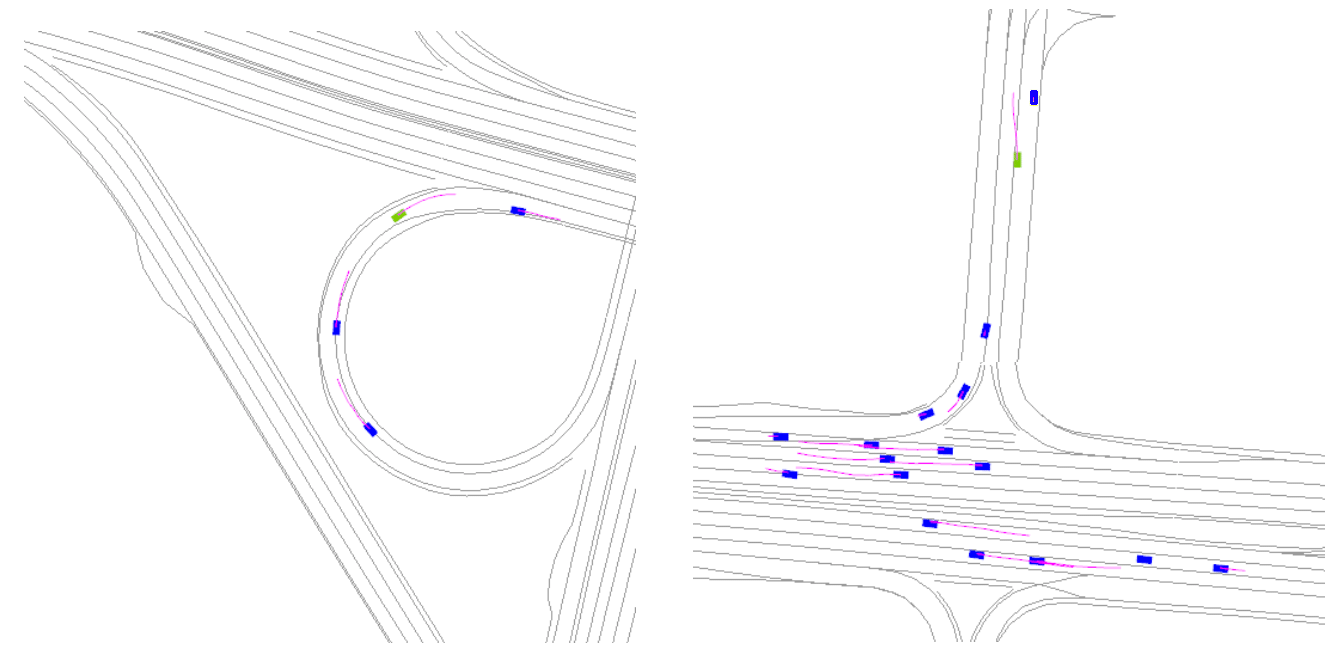}
\end{subfigure}
\caption{Traffic simulations of PredictionNet. Green rectangle denotes the ego vehicle, blue rectangles are other actors, and magenta lines denote predictions.  PredictionNet is able to generate realistic traffic under incomplete maps with varying topology.}
\label{fig:sim_offroute}
%\vspace{-4mm}
\end{figure}

After training on real-world data, PredictionNet can be used as a traffic simulation model. We initialize the state of traffic episodes with real data. Then we use closed-loop roll-outs, using the DNN as a stepping function, to simulate traffic offline. The network acts as a learned state-transition dynamics function that can be used to test planners in simulation or learn policies via reinforcement learning (RL).

We use the RNN's latent encoding at the current time as a proxy for the state $s_t = h_t$.
This latent state encodes the traffic history and lane geometry. %(Fig.~\ref{fig:Prednet_latent}).
Let $q_t$ be the configuration of the system, i.e., the positions and velocities of all actors.
Given the current action $a_t$, we rasterize $q_t$ and $a_t$ into the input $x_t$, i.e., then encoded into the latent state $s_t$.
We then rollout PredictionNet, and use post-processing to fit a unicycle kinematic model~\cite{lavalle2006planning} to the decoder output $y_{t+1}$.
This yields the configuration $q_{t+1}$ to forward the simulation and close the loop with the updated history.  
%We denote the transition function of PredictionNet by $s_{t+1}=f_{\text{traffic}}(s_t,a_t)$. 
Some examples are shown in Fig.~\ref{fig:sim_offroute}.
Note that whereas most systems provide only single-agent predictions, our DNN provides simultaneous multi-agent predictions conditioned on the ego-vehicle actions and other vehicle behaviors.

\subsection{Extending PredictionNet via Reinforcement Learning}
Since PredictionNet is trained with supervised learning, it can be seen as a type of imitation learning.
One drawback of imitation learning is its inability to handle rare/unsafe events that are not properly captured in the training data, such as cut-ins and harsh braking that might lead to collisions. 
To address these issues, we use a model-based RL framework to formulate the driving problem as an MDP, with PredictionNet as the transition model. We train an additional policy head to produce ego actions, and we call this extension \textit{PredictionNet-RL}.

% We formalize autonomous driving tasks as a Markov Decision Process, using the tuple: ${P}=\{{S},{A},{T},{R},{\gamma}\}$. In this formalization,  ${S}$ and ${A}$ denote the state and action space respectively. ${T}: {S}\times {A}\to {S}$ is the transition dynamics that PredictionNet will serve for and ${R}:{S} \times {A}\to \mathbb{R}$ is the task-specific reward function,  and $\gamma=[0,1]$ is a discount factor. Through this formalization, we generate diverse experience for the ego-actor for an improved best-response to potentially unsafe behaviors by surrounding actors. 

In this work, we focus primarily on avoiding cut-ins and collisions caused by harsh braking, both of which are challenges for traditional adaptive cruise control (ACC). Each episode consists of three cars: a lead car, a neighbor car, and the ego car. Initial states are randomized. Non-ego cars are controlled by the PredictionNet prediction head, while the ego car is controlled by the RL policy head. In the cut-in task, the neighbor car executes a cut-in move when it observes a gap between the lead car and the ego car. In the collision-avoidance task, the lead car executes a random harsh braking event.

Our goal is to learn a policy $\pi: s_t \mapsto a_t$ that maps the state $s_t$ at time $t$ to an action $a_t$, which is the ego vehicle's acceleration. The task reward is defined as $r_{\text{acc}}+ r_{\text{rare}}$ where $r_{\text{acc}}$ is a sum of manually designed reward terms such as distance to the lead car, acceleration changes, \emph{etc.}, and $r_{\text{rare}}$ is a sparse penalty term accounting for the rare events such as cut-ins and collisions. The trained RL policy learns to plan ahead, to reduce the gap and avoid cut-ins, and to keep a slower speed and avoid future harsh braking.

% We hypothesise that these rare long-horizon tasks are difficult for traditional Adaptive Cruise Control (ACC) policy and for Imitation Learning (IL) based policy as they often exhibit short-sighted behaviors, leaving excessive space for cut-ins and not decelerating early enough when cut-ins occur. As a remedy, RL policy can better learn to plan ahead, to reduce the gap and avoid cut-ins, or to keep a slower speed and avoid future harsh braking.

\section{EXPERIMENTS}

We investigate the following questions: (1) How well does PredictionNet perform on trajectory prediction benchmarks and internal datasets? (2) Can PredictionNet be used in a real-world actuated drive, and how does it compare to classical prediction methods? (3) Can we achieve feasible and reactive traffic simulation with PredictionNet? (4) Can we use PredictionNet-RL to improve ego vehicle behavior when encountering rare events?

\subsection{PredictionNet Performance}
%%%%%%%%%%%%%%%%%%%%%%%%%%%%%%%%%%%%%%%%%%%%%
% Architecture
 
\begin{table}
\caption{Network architecture, with ${\overline T}=6$ and $T=18$.  Timesteps are separated by 166.7~ms, so the network processes 1 second of past-present data, and predicts 3 seconds into the future.} 
%We use $15$ RNN cells at training time and $18$ RNN cells at inference time for future processing}
\begin{center}
\begin{minipage}[t]{.95\linewidth}
\resizebox{\linewidth}{!}{
\begin{tabular}{l|l|l|l}
	\textbf{Layer} & \textbf{Layer description} &	\textbf{Input} & \textbf{Output dimensions} \\
\hline 
	\hline
	\multicolumn{3}{c}{\textbf{Inputs:}}	 \\
0a & \emph{Input image (dynamic)}	& -- & $6 \times 3 \times 512 \times 512$  \\
0b & \emph{Input image (static)}	& -- & $1 \times 2 \times 512 \times 512$  \\
	\hline
	\multicolumn{3}{c}{\textbf{Encoder:}}	 \\
%	& \emph{2D feature extractors:}	 \\
%	\hline
1	& 2D convolution, ReLU	& 0a & $6 \times 16 \times 256 \times 256$ \\
2	& 2D convolution, ReLU	& 0b & $1 \times 16 \times 256 \times 256$ \\
2a	& Add &  1,2 & $6 \times 16 \times 256 \times 256$ \\
3	& ResNet block & 2a & $6 \times 32 \times 128 \times 128$ \\
4	& ResNet block & 3 & $6 \times 64 \times 64 \times 64$ \\
5	& ResNet block & 4 & $6 \times 64 \times 64 \times 64$ \\
	\hline
	\multicolumn{3}{c}{\textbf{RNN (past-present):}}	 \\
6--11	& 2D conv RNN, ASPPBlock\cite{chen2016deeplab}, ReLU &  5 & $64 \times 64 \times 64$ \\
	\hline
	\multicolumn{3}{c}{\textbf{RNN (future):}}	 \\
12--29	& 2D conv RNN, ASPPBlock\cite{chen2016deeplab}, ReLU &  11 & $64 \times 64 \times 64$ \\
	\hline
	\multicolumn{3}{c}{\textbf{Decoder:}}	 \\
30	& Transposed ResNet block &  29 & $24 \times 64 \times 64 \times 64$ \\
31	& Transposed ResNet block &  30 & $24 \times 32 \times 64 \times 64$ \\
32	& Transposed ResNet block &  31 & $24 \times 16 \times 64 \times 64$ \\
33	& Upsample via 2D convolution &  32 & $24 \times 5 \times 128 \times 128$ \\
34	& Upsample (only training) &  33 & $24 \times 5 \times 512 \times 512$
\end{tabular}
}\end{minipage}

\end{center}
\label{tab:netarch}
\vspace{-4mm}
\end{table}

We summarize the detailed network architecture in Table~\ref{tab:netarch}. The network is trained on $512 \times 512$ images as inputs and outputs with $0.33$ meter resolution per pixel thus providing $170 \times 170$~m top-down field of view (FoV). 
The latent space is $64 \times 64 \times 64$.
The focal loss weight $\lambda_0=0.05$, and other losses use $\lambda_1=\lambda_2=1.0$. 
%For the internal dataset, the FoV covers $160 \times 160$~m area, and each pixel corresponds to 0.31 meters. 
We compare our system with an analytical prediction baseline that utilizes kinematic models, assuming constant velocities and lane following. We use the following metrics to measure the prediction accuracy: Average Displacement Error (ADE) between corresponding points on predicted and ground truth trajectories, and Final Displacement Error (FDE) computed using only the final positions. Both are in meters.

% We use two main metrics to evaluate prediction accuracy: Average Displacement Error (ADE) and Final Displacement Error (FDE). ADE is average distance/displacement (in meters) between points on a predicted trajectory and corresponding points on a ground truth trajectory, whereas FDE is computed using only the final positions.
% We tested the accuracy of our model on the popular INTERACTION dataset \cite{zhan2019interaction} and on our own internal dataset. 

% We achieved competitive results on the INTERACTION dataset listed in Table \ref{tab:interaction_results}. This dataset was produced from drone's camera footage with manual top-down labeling. Our DNN was trained with 512x512 input and output resolution with 0.25 meter resolution per pixel. This provide 128x128 meters top-down field of view. The dataset map and past agent positions were rasterized as a top-down view and we used them for training. Unlike other systems we compared to, our DNN provides multi-agent predictions conditioned on the ego-vehicle actions all at once; most systems provide single-agent predictions. 

Our PredictionNet achieves competitive results on the INTERACTION dataset \cite{zhan2019interaction}, see Table \ref{tab:interaction_results}.
We ran the network on both the validation (V) set and the ``regular'' test set (R).  
The ``generalizability'' test set (G) was not available at the time of writing.
All published networks are listed.
While it is difficult to compare across dataset versions, we would like to stress that our network has demonstrated generalization across locations and sensor modalities in real-world driving.  As a result, we believe it is fair to compare our results with those on the generalizability (G) set, where our method is competitive.
Note that our system is the only one that is evaluated using a single trajectory (most other systems produce $6$ and measure the ADE/FDE of the best one), and ours is the only one that runs in real-time (RT).

\begin{figure}
    \centering
\begin{subfigure}{0.8\linewidth}
 \includegraphics[width=1\linewidth]{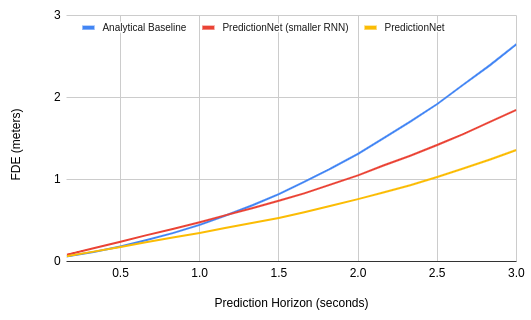}
\end{subfigure}%
\caption{FDE performances on internal dataset across horizons.}
\label{fig:prednet_internal_FDE}
%\vspace{-4mm}
\end{figure}

We also achieve good results on a large internal driving dataset produced from LiDAR perception data, see Table~\ref{tab:internal_results}.  
The dataset is a mix of freeway and urban scenarios with both light and heavy traffic, containing more than $5000$~km.
Our approach outperforms an analytical baseline that relies on lane following and constant speed heuristics. 
Fig.~\ref{fig:prednet_internal_FDE} shows that
the analytical approach diverges considerably beyond $0.7$ seconds, and that reducing the size of the latent space to $64 \times 32 \times 32$ improves inference time by $30\%$ but yields larger errors.

\begin{table}
\caption{Comparison on INTERACTION dataset \cite{zhan2019interaction}. Results were taken from INTERPRET Challenge, NEURIPS20 stage, at 3 seconds prediction horizon.  See text for details.
}
\centering
\begin{tabular}{c|cccll}
\backslashbox{Model}{Metric} & set & 1-traj & RT & ADE ($\downarrow$) & FDE ($\downarrow$) \\
\hline
DESIRE~\cite{lee2017desire} & V & \redxmark & \redxmark & $0.32$  & $0.88$ \\
MultiPath~\cite{chai2019multipath} & V & \redxmark & \redxmark & $0.30$  & $0.99$ \\
TNT \cite{zhao2020tnt} & V & \redxmark & \redxmark & $0.21$  & $0.67$ \\
ReCoG~\cite{mo2021:recog} & V & \redxmark & \redxmark & $0.192$  & $0.646$ \\
ITRA~\cite{scibior2021:itra} & V & \redxmark & \redxmark & $0.17$  & $0.49$ \\
PredictionNet (ours) & V & \greencheckmark & \greencheckmark & $0.458$  & $1.030$ \\
%ReCog & T & \redxmark & \redxmark & $0.192$  & $0.646$ \\
PredictionNet (ours) & R & \greencheckmark & \greencheckmark & $0.518$  & $1.228$ \\
SimNet \cite{bergamini2021simnet} & G & \redxmark & \redxmark & $0.652$  & $1.670$ \\
MIFNet & G & \redxmark & \redxmark & $0.534$ & $1.425$ \\
\end{tabular}
\label{tab:interaction_results}
%\vspace{-4mm}
\end{table}

\def\greencheckmark{\textcolor{darkgreen}{\checkmark}}
\def\redxmark{\textcolor{darkred}{\ding{55}}}  % pifont

%as of September 7, 2021 and listed for recognized networks. 

% DO NOT REMOVE THIS TABLE! But we won't use it for now, but we might in the future
% \begin{table}
% \centering
% \begin{tabular}{c|cc}
% method & min ADE & min FDE \\
% \hline
% multipath++ [?] & 0.25 & 0.48 \\
% Tsinghua MARS DenseTNT [?] & 0.39 & 0.70 \\
% ours & 1.07 & 2.08 \\
% \end{tabular}
% \caption{Experimental results on Motion Prediction Challenge (Waymo Open Dataset) at 3 seconds prediction horizon. ADE and FDE metrics are listed in meters.}
% \label{tab:waymo_results}
% \vspace{-4mm}
% \end{table}
 
\begin{table}
\caption{Comparison on internal dataset at 3 seconds horizon.}
\centering
\begin{minipage}[t]{.8\linewidth}
\resizebox{\linewidth}{!}{
\begin{tabular}{c|cc}
\backslashbox{Model}{Metric} & ADE ($m$) & FDE ($m$) \\
\hline
analytical baseline & $1.10$ & $2.65$ \\
PredictionNet (ours) & $\mathbf{0.62}$ & $\mathbf{1.36}$ \\
\end{tabular}}
\end{minipage}
\label{tab:internal_results}
%\vspace{-7mm}
\end{table}

\subsection{Real-world Autonomous Driving}
 
We integrated our PredictionNet system into an actuated real-world vehicle. We use the DNN to predict 3 seconds into the future, and then extrapolate the result from $3$ to $5$ seconds using the analytical predictor. We chose this combination as a trade-off between accuracy and performance: the DNN reacts to important events within the $3$ second horizon, while analytical extrapolation reduces run-time latency.

We assume that better prediction leads to better planning and hence more driving comfort by avoiding harsh braking or emergency disengagements. 
To quantify performance, each route is split into segments of a few seconds each, and the average jerk (derivative of acceleration) is computed for each segment. We classify segments as ``comfortable'' or ``not comfortable'' based on a threshold from trials with human drivers. 
The final score is the percentage of route segments that are comfortable. 

This system was used for actuated drives on routes in Washington state and California for approximately $530$~km in total. 
The network consumes perception output from cameras and radars, even though it was trained on LiDAR-derived data. 
We encountered no network-related safety disengagements. The DNN was able to handle non-mapped areas in the real world thanks to a training procedure that includes random map dropouts. Compared to the analytic baseline, PredictionNet improved the mean comfort score by \textbf{15\%} and reduced standard deviation by \textbf{18\%}. 

PredictionNet is also computationally efficient. The DNN inference time is around \textbf{5~ms} on an embedded NVIDIA's Drive AGX GPU and \textbf{4~ms} on a Titan RTX GPU. The pre-/post-processing time is \textbf{3~ms} and  \textbf{2~ms}, respectively. To our knowledge, our system is the fastest DNN-based approach available for real-world traffic prediction.
By comparison, our closest competitor MultiPath~\cite{chai2019multipath} is an order of magnitude slower, and
Trajectron++~\cite{salzmann2020trajectron++} is two orders of magnitude slower.

% although run-time computation time is hard to discern for existing system. Since most papers do not provide it

% Run-time speed performance data is hard to come by for prediction systems, since most papers do not provide it. SimNet \cite{bergamini2021simnet} suggests that ``During evaluation, the whole pipeline takes around 400ms per frame with a modern GPU...''.  In comparison, our model runs at \textbf{5 ms DNN inference plus 3 ms pre-/post-processing time} on the embedded NVIDIA's Drive AGX computer in vehicle and can be used for real-world autonomous driving and planning with multiple roll-outs.

\subsection{Simulation Performance}

To measure the adequacy of PredictionNet for simulation, we used it as an environment step function of a standard OpenAI Gym environment \cite{brockman2016openai}. We used our large internal dataset containing dense multi-agent interactions to sample initial traffic states. To measure performance, we focused on collision rates and off-route driving rates, which were computed using the environment's contact detectors and the lane geometry (to determine non-drivable areas), respectively. 

Table \ref{tab:model_sim} shows the simulation failure rates among all environment steps for both PredictionNet and an analytical baseline. 
Log-replay is applied for $1.33$~seconds, then all vehicles are controlled for an additional $6.67$ seconds by either PredictionNet or 
the unicycle model \cite{lavalle2006planning}, the latter of which is applied to vehicle kinematics while maintaining constant heading and speed.
Total episode length is $8.0$~seconds.
Failure is declared when the vehicle leaves the drivable area (``off-road'') or intersects another vehicle (``collision'').
Furthermore, for ``reactive collision'' we apply harsh braking to random agents. 
We observe that PredictionNet reliably avoids collisions and remains in the drivable area. Our approach outperforms the baseline on all metrics by a large margin. 

\begin{table}
\caption{Simulation comparison, showing the percentage of failures over all environment steps.}
\centering
\begin{minipage}[t]{.9\linewidth}
\resizebox{\linewidth}{!}
{\begin{tabular}{c|ccc}
\multicolumn{0}{c|}{\backslashbox{Model}{Metric}}   &
  \multicolumn{0}{c}{off-road} & \multicolumn{0}{c}{collision} & \multicolumn{0}{c}{reactive collision}  \\ \hline
baseline & 2.1$\%$  &  8.7$\%$ &  7.1$\%$ \\
PredictionNet &   $\mathbf{1.4\%}$  &  $\mathbf{4.1\%}$  &  $\mathbf{5.3\%}$ \\
\end{tabular}}
\end{minipage}
\label{tab:model_sim}
%\vspace{-4mm}
\end{table}

\subsection{Ego-Policy Performance}
\begin{figure}
    \centering
\begin{subfigure}{1\linewidth}
 \includegraphics[width=1.0\linewidth]{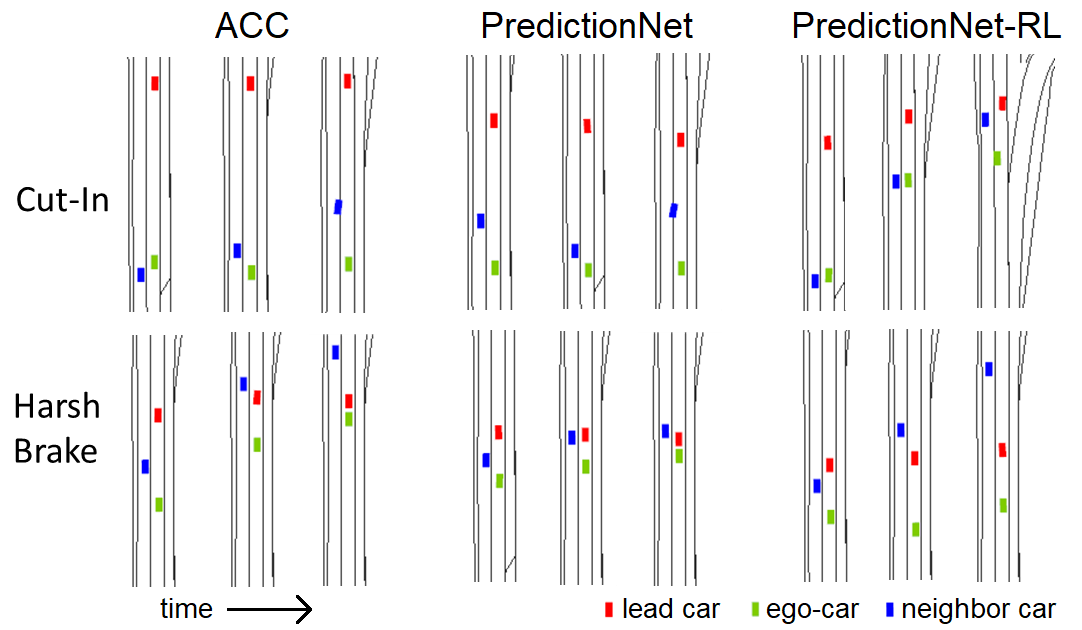}
\end{subfigure}
\caption{PredictionNet-RL outperforms both PredictionNet and ACC for the cut-in and harsh braking scenarios. Shown are 3 time slices from left to right for each policy.}
\label{fig:policy_brake}
\vspace{-4mm}
\end{figure}

We use reinforcement learning (RL) to 
improve handling of rare and unsafe events like cut-ins and harsh-braking.
We aim to improve upon both traditional adaptive cruise control (ACC) and imitation learning-based driving policies.  Our PredictionNet-RL uses the Soft Actor Critic algorithm (SAC) \cite{haarnoja2018soft} to train the policy and critic heads, while keeping PredictionNet trunk weights frozen. 
The latent vector $s_t=h_t$ is fed into three fully convolutional layers to extract features, and each of the policy and critic heads is a three-layer MLP. The MDP discount factor is $\gamma=0.95$. The ACC reward function is $r_{\text{acc}}=\alpha_1 |\Delta_a| +\alpha_2 |\Delta_v| +\alpha_3 d$  where $\Delta_a,\Delta_v$ denote the changes in acceleration and velocity, and $d$ denotes the distance to the lead car; $\alpha_{1,2,3}$ are tuned to accomplish smooth following in real-world scenarios.  We use episodes with randomized initial conditions. An episode ends if a collision or cut-in happens with task-specific penalty $r_{\text{rare}}=-1$, or reaches the maximum number ($48$) of episode steps. We generate and detect cut-ins or braking signals based on agents' headings and accelerations. 
 
We evaluate our learned policies in separate cut-in and harsh-braking tasks. We compare PredictionNet-RL policy with vanilla PredictionNet ego-prediction that only captures expert driving via supervised learning. We also compare with an internal implementation of ACC.
%that is trained without rare event rewards and no PredictionNet input, which can be seen as a hand-engineered planner that handles general cases. 
In Table \ref{tab:model_policy}, we test each policy with $10$ different runs, observing that PredictionNet-RL reaches the best performance for both tasks. 

Fig. \ref{fig:policy_brake} shows some qualitative results.
In the cut-in scenarios, the neighbor car (blue) attempts to cut in between the lead car (red) and ego car (green). Whereas both ACC and vanilla PredictionNet allow this cut-in,
PredictionNet-RL learns to speed up to prevent these aggressive cut-in maneuvers.  
In the harsh braking case, the lead car (red) performs sudden braking.
Whereas both ACC and PredictionNet cause the ego car to come dangerously close to the lead car, the PredictionNet-RL agent learns to slow down to avoid collision with the lead car. Overall, such sparse-reward scenarios are challenging for the baselines since the rare events are underrepresented in the training data. 

\begin{figure}
    \centering
\begin{subfigure}{0.90\linewidth}
 \includegraphics[width=1\linewidth]{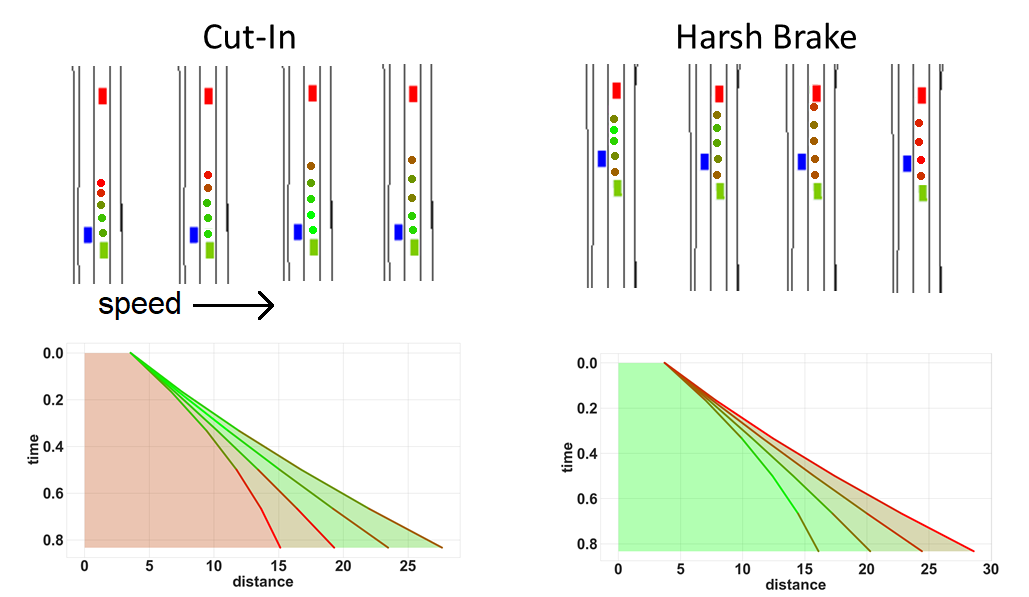}
\end{subfigure}%
\caption{Visualization of the RL value function for $4$ speed profiles for each of the two tasks. {\sc Top:} Top-down view, with circles denoting sampled states colored by the value function. {\sc Bottom:} Value function computed along $4$ speed profiles. The $y$-axis denotes the time horizon, while $x$-axis denotes the distance from the state $s_{t+1}$. Each trajectory curve is colored according to the value function, with green denoting higher critic rewards, and red denoting lower rewards. The space between speed profile curves is filled with the average nearby color.}
\label{fig:acc_value_fn}
\vspace{-4mm}
\end{figure}

Fig. \ref{fig:acc_value_fn} visualizes the learned RL value function~\cite{Sutton1998book} for $4$ possible speed profiles for each of the two tasks, together with sampled top-down timesteps.
As traffic speed increases, the ego-vehicle is rewarded for speeding up to discourage cut-in, and for braking quickly to avoid collision.

\begin{table}
\caption{Policy learning comparison, showing the  percentage of failures for $10$ test episodes.}
\centering
\begin{minipage}[t]{.9\linewidth}
\resizebox{\linewidth}{!}
{\begin{tabular}{c|ccc}
\multicolumn{0}{c|}{\backslashbox{Task}{Policy}}  & \multicolumn{0}{c}{ACC} &
  \multicolumn{0}{c}{PredictionNet} & \multicolumn{0}{c}{PredictionNet-RL}  \\ \hline
Cut-In &   88$\%$ & 56$\%$  &  $\mathbf{2}\%$ \\
Harsh Brake &   100$\%$  &  60$\%$  &   $\mathbf{0}\%$  \\
\end{tabular}}
\end{minipage}
\label{tab:model_policy}
%\vspace{-4mm}
\end{table}

%\subsection{Ego-Policy Ablation Study}
%Since PredictionNet provides the world model for simulating agent roll-outs and the latent states for the Q-function, we can sample rollouts and evaluate the scores of the sampled states. Specifically to interpret the learned value functions, we use the latent states along the predicted trajectory from PredictionNet. We then sample ego-agent action sequences and  use the critic to compute the average state action values.  Fig. \ref{fig:acc_value_fn} presents the speed-profile view aligned with the top-down projection view for the value functions. For the cut-ins, we observe a balance between avoiding cut-in and collision by giving high value for maintaining a gap and discouraging speeding up or slowing down. For harsh-braking, slowing down while maintaining comfort is preferred when the ego car is close to the leading car. The state value estimation is useful for trajectory optimization methods such as model predictive control. 

\section{CONCLUSION}
In this work, we presented PredictionNet, a real-time DNN for trajectory prediction that can efficiently forecast trajectories of all traffic actors simultaneously. We showed that a well-trained prediction system provides an efficient data-driven traffic simulator that significantly outperforms heuristic-based baselines. Furthermore, our experiments with an RL-based extension show significant improvement for rare events that are not captured in the expert dataset. We validated our approach in simulation, on off-line datasets, and via on-line real-world actuated driving. For future work, we plan to incorporate the direct trajectory regression into the DNN, use live map perception, and transfer our RL experiments to the real vehicle.

\section*{Acknowledgment}

We would like to acknowledge Rotem Aviv and Ruchi Bhargava for support. We thank Fangkai Yang and Ryan Oldja for system work.

% Simulations have the potentials to expedite the verification and training procedures for autonomous driving stacks. Existing simulators typically fall into two categories. The first one drives traffic through hand-coded rules and often fails to capture complex maneuvers and interactions such as cut-in \cite{dosovitskiy2017carla,lopez2018microscopic}. The second one replays passive observations from structured real-world driving logs and cannot be used to study counter-factual behaviors. We focus on applying the traffic prediction network for the simulation step \cite{bergamini2021simnet,suo2021trafficsim}, which controls the motion dynamics for traffic agents. We show that the closed-loop simulated traffics capture collision-free and lane-following behaviors, as well as allow interactive inspections for agent behaviors. A closely related concept is behavior cloning \cite{ross2011reduction}, since the traffic forecasting network output for each agent can be treated as policy learned from human demonstration data. Under a model-based RL framework, our method proposes to generate realistic, reactive traffics to train policies for long-horizon tasks that are rare in real-world data. We show that for harsh-braking and cut-in tasks the learned policy can outperform imitation learning agents and traditional controllers. The learned policy can potentially be used to remedy trajectory scoring function in the planning modules.
% end misc notes

% not sure about this
\clearpage
\bibliographystyle{IEEEtran}
\bibliography{PredictionNet}

\begin{thebibliography}{10}
\providecommand{\url}[1]{#1}
\csname url@rmstyle\endcsname
\providecommand{\newblock}{\relax}
\providecommand{\bibinfo}[2]{#2}
\providecommand\BIBentrySTDinterwordspacing{\spaceskip=0pt\relax}
\providecommand\BIBentryALTinterwordstretchfactor{4}
\providecommand\BIBentryALTinterwordspacing{\spaceskip=\fontdimen2\font plus
\BIBentryALTinterwordstretchfactor\fontdimen3\font minus
  \fontdimen4\font\relax}
\providecommand\BIBforeignlanguage[2]{{%
\expandafter\ifx\csname l@#1\endcsname\relax
\typeout{** WARNING: IEEEtran.bst: No hyphenation pattern has been}%
\typeout{** loaded for the language `#1'. Using the pattern for}%
\typeout{** the default language instead.}%
\else
\language=\csname l@#1\endcsname
\fi
#2}}

\bibitem{yoshida2008game}
W.~Yoshida, R.~J. Dolan, and K.~J. Friston, ``Game theory of mind,'' \emph{PLoS
  computational biology}, vol.~4, no.~12, 2008.

\bibitem{hernandez2019agent}
P.~Hernandez-Leal, B.~Kartal, and M.~E. Taylor, ``Agent modeling as auxiliary
  task for deep reinforcement learning,'' in \emph{Proceedings of the AAAI
  Conference on Artificial Intelligence and Interactive Digital Entertainment},
  vol.~15, no.~1, 2019, pp. 31--37.

\bibitem{lee2017desire}
N.~Lee, W.~Choi, P.~Vernaza, C.~B. Choy, P.~H. Torr, and M.~Chandraker,
  ``{DESIRE}: Distant future prediction in dynamic scenes with interacting
  agents,'' in \emph{Proceedings of the IEEE Conference on Computer Vision and
  Pattern Recognition (CVPR)}, 2017, pp. 336--345.

\bibitem{rhinehart2019precog}
N.~Rhinehart, R.~McAllister, K.~Kitani, and S.~Levine, ``{PRECOG}: {PRE}diction
  conditioned on goals in visual multi-agent settings,'' in \emph{Proceedings
  of the IEEE/CVF International Conference on Computer Vision (ICCV)}, 2019,
  pp. 2821--2830.

\bibitem{chai2019multipath}
Y.~Chai, B.~Sapp, M.~Bansal, and D.~Anguelov, ``{MultiPath}: Multiple
  probabilistic anchor trajectory hypotheses for behavior prediction,'' in
  \emph{CoRL}, 2019.

\bibitem{djuric2020multixnet}
N.~Djuric, H.~Cui, Z.~Su, S.~Wu, H.~Wang, F.-C. Chou, L.~S. Martin, S.~Feng,
  R.~Hu, Y.~Xu, \emph{et~al.}, ``{MultiXNet}: Multiclass multistage multimodal
  motion prediction,'' in \emph{IEEE Intelligent Vehicles Symposium}, 2020.

\bibitem{casas2020implicit}
S.~Casas, C.~Gulino, S.~Suo, K.~Luo, R.~Liao, and R.~Urtasun, ``Implicit latent
  variable model for scene-consistent motion forecasting,'' in \emph{ECCV},
  2020, pp. 624--641.

\bibitem{salzmann2020trajectron++}
T.~Salzmann, B.~Ivanovic, P.~Chakravarty, and M.~Pavone, ``Trajectron++:
  Dynamically-feasible trajectory forecasting with heterogeneous data,'' in
  \emph{ECCV}, 2020, pp. 683--700.

\bibitem{zhan2019interaction}
W.~Zhan, L.~Sun, D.~Wang, H.~Shi, A.~Clausse, M.~Naumann, J.~Kummerle,
  H.~Konigshof, C.~Stiller, A.~de~La~Fortelle, \emph{et~al.}, ``Interaction
  dataset: An international, adversarial and cooperative motion dataset in
  interactive driving scenarios with semantic maps,'' \emph{arXiv preprint
  arXiv:1910.03088}, 2019.

\bibitem{kingma2013auto}
D.~P. Kingma and M.~Welling, ``Auto-encoding variational {B}ayes,'' \emph{arXiv
  preprint arXiv:1312.6114}, 2013.

\bibitem{gao2020vectornet}
J.~Gao, C.~Sun, H.~Zhao, Y.~Shen, D.~Anguelov, C.~Li, and C.~Schmid,
  ``{VectorNet}: Encoding {HD} maps and agent dynamics from vectorized
  representation,'' in \emph{Proceedings of the IEEE/CVF Conference on Computer
  Vision and Pattern Recognition (CVPR)}, 2020.

\bibitem{wu2020comprehensive}
Z.~Wu, S.~Pan, F.~Chen, G.~Long, C.~Zhang, and S.~Y. Philip, ``A comprehensive
  survey on graph neural networks,'' \emph{IEEE transactions on neural networks
  and learning systems}, vol.~32, no.~1, pp. 4--24, 2020.

\bibitem{dosovitskiy2017carla}
A.~Dosovitskiy, G.~Ros, F.~Codevilla, A.~Lopez, and V.~Koltun, ``{CARLA}: An
  open urban driving simulator,'' in \emph{Conference on robot learning
  (CoRL)}, 2017.

\bibitem{lopez2018microscopic}
P.~A. Lopez, M.~Behrisch, L.~Bieker-Walz, J.~Erdmann, Y.-P. Fl{\"o}tter{\"o}d,
  R.~Hilbrich, L.~L{\"u}cken, J.~Rummel, P.~Wagner, and E.~Wie{\ss}ner,
  ``Microscopic traffic simulation using {SUMO},'' in \emph{International
  Conference on Intelligent Transportation Systems (ITSC)}, 2018, pp.
  2575--2582.

\bibitem{suo2021trafficsim}
S.~Suo, S.~Regalado, S.~Casas, and R.~Urtasun, ``{TrafficSim}: Learning to
  simulate realistic multi-agent behaviors,'' in \emph{Proceedings of the
  IEEE/CVF Conference on Computer Vision and Pattern Recognition (CVPR)}, 2021.

\bibitem{bergamini2021simnet}
L.~Bergamini, Y.~Ye, O.~Scheel, L.~Chen, C.~Hu, L.~Del~Pero, B.~Osinski,
  H.~Grimmett, and P.~Ondruska, ``{SimNet}: Learning reactive self-driving
  simulations from real-world observations,'' in \emph{ICRA}, 2021.

\bibitem{buehler2009darpa}
M.~Buehler, K.~Iagnemma, and S.~Singh, \emph{The DARPA Urban Challenge:
  Autonomous Vehicles in City Traffic}.\hskip 1em plus 0.5em minus 0.4em\relax
  Springer, 2009, vol.~56.

\bibitem{fan2018baidu}
H.~Fan, F.~Zhu, C.~Liu, L.~Zhang, L.~Zhuang, D.~Li, W.~Zhu, J.~Hu, H.~Li, and
  Q.~Kong, ``Baidu {A}pollo {EM} motion planner,'' in \emph{arXiv:1807.08048},
  2018.

\bibitem{zeng2019end}
W.~Zeng, W.~Luo, S.~Suo, A.~Sadat, B.~Yang, S.~Casas, and R.~Urtasun,
  ``End-to-end interpretable neural motion planner,'' in \emph{Proceedings of
  the IEEE/CVF Conference on Computer Vision and Pattern Recognition}, 2019,
  pp. 8660--8669.

\bibitem{sadat2019jointly}
A.~Sadat, M.~Ren, A.~Pokrovsky, Y.-C. Lin, E.~Yumer, and R.~Urtasun, ``Jointly
  learnable behavior and trajectory planning for self-driving vehicles,'' in
  \emph{IEEE/RSJ International Conference on Intelligent Robots and Systems
  (IROS)}, 2019, pp. 3949--3956.

\bibitem{henaff2019model}
M.~Henaff, A.~Canziani, and Y.~LeCun, ``Model-predictive policy learning with
  uncertainty regularization for driving in dense traffic,'' in \emph{ICLR},
  2019.

\bibitem{pomerleau1989alvinn}
D.~A. Pomerleau, ``{ALVINN}: An autonomous land vehicle in a neural network,''
  in \emph{Proceedings of Neural Information Processing Systems (NeurIPS)},
  1989.

\bibitem{ratliff2006maximum}
N.~D. Ratliff, J.~A. Bagnell, and M.~A. Zinkevich, ``Maximum margin planning,''
  in \emph{Proceedings of the 23rd International Conference on Machine Learning
  (ICML)}, 2006, pp. 729--736.

\bibitem{bansal2018chauffeurnet}
M.~Bansal, A.~Krizhevsky, and A.~Ogale, ``{ChauffeurNet}: Learning to drive by
  imitating the best and synthesizing the worst,'' in \emph{Robotics Science
  and Systems (RSS)}, 2019.

\bibitem{codevilla2018end}
F.~Codevilla, M.~M{\"u}ller, A.~L{\'o}pez, V.~Koltun, and A.~Dosovitskiy,
  ``End-to-end driving via conditional imitation learning,'' in \emph{IEEE
  International Conference on Robotics and Automation (ICRA)}, 2018, pp.
  4693--4700.

\bibitem{ross2011reduction}
S.~Ross, G.~Gordon, and D.~Bagnell, ``A reduction of imitation learning and
  structured prediction to no-regret online learning,'' in \emph{Proceedings of
  the Fourteenth International Conference on Artificial Intelligence and
  Statistics}, 2011, pp. 627--635.

\bibitem{wolf2017learning}
P.~Wolf, C.~Hubschneider, M.~Weber, A.~Bauer, J.~H{\"a}rtl, F.~D{\"u}rr, and
  J.~M. Z{\"o}llner, ``Learning how to drive in a real world simulation with
  deep {Q}-networks,'' in \emph{IEEE Intelligent Vehicles Symposium (IV)},
  2017, pp. 244--250.

\bibitem{kiran2021deep}
B.~R. Kiran, I.~Sobh, V.~Talpaert, P.~Mannion, A.~A. Al~Sallab, S.~Yogamani,
  and P.~P{\'e}rez, ``Deep reinforcement learning for autonomous driving: A
  survey,'' \emph{IEEE Transactions on Intelligent Transportation Systems},
  2021.

\bibitem{shalev2016safe}
S.~Shalev-Shwartz, S.~Shammah, and A.~Shashua, ``Safe, multi-agent,
  reinforcement learning for autonomous driving,'' in \emph{NeurIPS workshop},
  2016.

\bibitem{lillicrap2015continuous}
T.~P. Lillicrap, J.~J. Hunt, A.~Pritzel, N.~Heess, T.~Erez, Y.~Tassa,
  D.~Silver, and D.~Wierstra, ``Continuous control with deep reinforcement
  learning,'' in \emph{ICLR}, 2015.

\bibitem{chen2019model}
J.~Chen, B.~Yuan, and M.~Tomizuka, ``Model-free deep reinforcement learning for
  urban autonomous driving,'' in \emph{IEEE Intelligent Transportation Systems
  Conference (ITSC)}, 2019, pp. 2765--2771.

\bibitem{saxena2020driving}
D.~M. Saxena, S.~Bae, A.~Nakhaei, K.~Fujimura, and M.~Likhachev, ``Driving in
  dense traffic with model-free reinforcement learning,'' in \emph{IEEE
  International Conference on Robotics and Automation (ICRA)}, 2020, pp.
  5385--5392.

\bibitem{lin2017focal}
T.-Y. Lin, P.~Goyal, R.~B. Girshick, K.~He, and P.~Dollar, ``Focal loss for
  dense object detection,'' in \emph{ICCV}, 2017.

\bibitem{lavalle2006planning}
S.~M. LaValle, \emph{Planning Algorithms.}\hskip 1em plus 0.5em minus
  0.4em\relax Cambridge University Press, 2006.

\bibitem{chen2016deeplab}
L.~Chen, G.~Papandreou, I.~Kokkinos, K.~Murphy, and A.~L. Yuille, ``{DeepLab}:
  Semantic image segmentation with deep convolutional nets, atrous convolution,
  and fully connected {CRF}s,'' \emph{IEEE Transactions on Pattern Analysis and
  Machine Intelligence (PAMI)}, vol.~40, no.~4, Apr. 2018.

\bibitem{zhao2020tnt}
H.~Zhao, J.~Gao, T.~Lan, C.~Sun, B.~Sapp, B.~Varadarajan, Y.~Shen, Y.~Shen,
  Y.~Chai, C.~Schmid, C.~Li, and D.~Anguelov, ``{TNT:} target-driven trajectory
  prediction,'' in \emph{arXiv:2008.08294}, 2020.

\bibitem{mo2021:recog}
X.~Mo, Y.~Xing, and C.~Lv, ``{ReCoG}: A deep learning framework with
  heterogeneous graph for interaction-aware trajectory prediction,'' in
  \emph{arXiv:2012.05032}, 2021.

\bibitem{scibior2021:itra}
A.~Scibior, V.~Lioutas, D.~Reda, P.~Bateni, and F.~Wood, ``Imagining the road
  ahead: Multi-agent trajectory prediction via differentiable simulation,'' in
  \emph{IEEE International Conference on Intelligent Transportation Systems
  (ITSC)}, 2021.

\bibitem{brockman2016openai}
G.~Brockman, V.~Cheung, L.~Pettersson, J.~Schneider, J.~Schulman, J.~Tang, and
  W.~Zaremba, ``{OpenAI} gym,'' \emph{arXiv preprint arXiv:1606.01540}, 2016.

\bibitem{haarnoja2018soft}
T.~Haarnoja, A.~Zhou, P.~Abbeel, and S.~Levine, ``Soft actor-critic: Off-policy
  maximum entropy deep reinforcement learning with a stochastic actor,'' in
  \emph{International Conference on Machine Learning (ICML)}, 2018, pp.
  1861--1870.

\bibitem{Sutton1998book}
R.~S. Sutton and A.~G. Barto, \emph{Reinforcement Learning: An Introduction},
  2nd~ed.\hskip 1em plus 0.5em minus 0.4em\relax The MIT Press, 2018.

\end{thebibliography}

\end{document}